\documentclass{article}
\usepackage{spconf,amsmath,graphicx}

\usepackage{hyperref}       
\usepackage{booktabs}       
\usepackage{xspace}         
\usepackage{bm}             
\usepackage{amssymb}        
\usepackage{pifont}         

\newcommand{\cmark}{\ding{51}}
\newcommand{\xmark}{-}

\newcommand{\PM}{$\pm$\xspace}

\newcommand{\s}{\bm{s}}
\renewcommand{\d}{\bm{d}}
\renewcommand{\c}{\bm{c}}
\newcommand{\y}{\bm{y}}
\newcommand{\h}{\bm{h}}
\newcommand{\f}{\bm{f}}
\newcommand{\g}{\bm{g}}
\newcommand{\p}{\bm{p}}
\renewcommand{\b}{\bm{b}}
\newcommand{\x}{\bm{x}}

\newcommand{\wb}{\bm{w}}
\newcommand{\mub}{\bm{\mu}}
\newcommand{\sigmab}{\bm{\sigma}}
\newcommand{\Zb}{\bm{Z}}
\newcommand{\Db}{\bm{\Delta}}

\newcommand{\wh}{\hat{\wb}}

\newcommand{\sigmah}{\hat{\sigmab}}
\newcommand{\Dh}{\hat{\Db}}

\newcommand{\softmax}{\textrm{S}_\textrm{max}}
\newcommand{\e}[1]{\exp(#1)}
\newcommand{\softplus}{\textrm{S}_+}

\newcommand{\alphab}{\bm{\alpha}}
\newcommand{\eb}{\bm{e}}

\newcommand{\mcG}{\mathcal{G}}

\newcommand{\weblink}{https://google.github.io/tacotron/publications/location_relative_attention}

\title{Location-Relative Attention Mechanisms For Robust Long-Form \\ Speech Synthesis}
\name{
\begin{tabular}{c}
Eric Battenberg \quad
RJ Skerry-Ryan \quad
Soroosh Mariooryad \quad
Daisy Stanton \quad
\\
David Kao \quad
Matt Shannon \quad
Tom Bagby
\end{tabular}
}
\address{Google Research}

\begin{document}
\ninept
\maketitle
\begin{abstract}
Despite the ability to produce human-level speech for in-domain text, attention-based end-to-end text-to-speech (TTS) systems suffer from text alignment failures that increase in frequency for out-of-domain text.
We show that these failures can be addressed using simple location-relative attention mechanisms that do away with content-based query/key comparisons.
We compare two families of attention mechanisms: location-relative GMM-based mechanisms and additive energy-based mechanisms.
We suggest simple modifications to GMM-based attention that allow it to align quickly and consistently during training, and introduce a new location-relative attention mechanism to the additive energy-based family, called Dynamic Convolution Attention (DCA).
We compare the various mechanisms in terms of alignment speed and consistency during training, naturalness, and ability to generalize to long utterances,
and conclude that GMM attention and DCA can generalize to very long utterances, while preserving naturalness for shorter, in-domain utterances.

\end{abstract}

\section{Introduction}
\label{sec:introduction}

Sequence-to-sequence models that use an attention mechanism to align the input and output sequences~\cite{Graves:2013ua,Bahdanau:2014vz} are currently the predominant paradigm in end-to-end TTS.
Approaches based on the seminal Tacotron system~\cite{Wang:2017uz} have demonstrated naturalness that rivals that of human speech for certain domains~\cite{Shen:2017vg}.
Despite these successes, there are sometimes complaints of a lack of robustness in the alignment procedure that leads to missing or repeating words, incomplete synthesis, or an inability to generalize to longer utterances~\cite{Zhang:2018is,He:2019tg,Liu:2019us}.

The original Tacotron system~\cite{Wang:2017uz}
used the content-based attention mechanism introduced in \cite{Bahdanau:2014vz} to align the target text with the output spectrogram.
This mechanism is purely content-based and does not exploit the monotonicity and locality properties of TTS alignment, making it one of the least stable choices.
The Tacotron 2 system~\cite{Shen:2017vg} used the improved hybrid location-sensitive mechanism from \cite{Chorowski:2015uh} that combines content-based and location-based features, allowing generalization to utterances longer than those seen during training.

The hybrid mechanism still has occasional alignment issues which led a number of authors to develop attention mechanisms that directly exploit monotonicity~\cite{Raffel:2017vl,Zhang:2018is,He:2019tg}.
These monotonic alignment mechanisms have demonstrated properties like increased alignment speed during training, improved stability, enhanced naturalness, and a virtual elimination of synthesis errors. 
Downsides of these methods include decreased efficiency due to a reliance on recursion to marginalize over possible alignments, the necessity of training hacks to ensure learning doesn't stall or become unstable, and decreased quality when operating in a more efficient hard alignment mode during inference.

Separately, some authors~\cite{SkerryRyan:2018ub} have moved back toward the purely location-based GMM attention introduced by Graves in \cite{Graves:2013ua}, and some have proposed stabilizing GMM attention by using softplus nonlinearities in place of the exponential function~\cite{Kastner:2018wg,Battenberg:2019tc}.
However, there has been no systematic comparison of these design choices.

In this paper, we compare the content-based and location-sensitive mechanisms used in Tacotron 1 and 2 with a variety of simple location-relative mechanisms in terms of alignment speed and consistency, naturalness of the synthesized speech, and ability to generalize to long utterances.
We show that GMM-based mechanisms are able to generalize to very long (potentially infinite-length) utterances,
and we introduce simple modifications that result in improved speed and consistency of alignment during training.
We also introduce a new location-relative mechanism called Dynamic Convolution Attention that modifies the hybrid location-sensitive mechanism from Tacotron 2 to be purely location-based, allowing it to generalize to very long utterances as well.

\section{Two Families of Attention Mechanisms}
\label{sec:attention_mechanisms}

\subsection{Basic Setup}
\label{subsec:basic_setup}

The system that we use in this paper is based on the original Tacotron system~\cite{Wang:2017uz} with architectural modifications from the baseline model detailed in the appendix of \cite{Battenberg:2019tc}.
We use the CBHG encoder from \cite{Wang:2017uz} to produce a sequence of encoder outputs, $\{\h_j\}_{j=1}^L$, from a length-$L$ input sequence of target phonemes, $\{\x_j\}_{j=1}^L$. 
Then an attention RNN, \eqref{eq:attention_rnn}, produces a sequence of states, $\{\s_i\}_{i=1}^T$, that the attention mechanism uses to compute $\alphab_i$, the alignment at decoder step $i$.  
Additional arguments to the attention function in \eqref{eq:attention_fn_context_vector} depend on the specific attention mechanism (e.g., whether it is content-based, location-based, or both).
The context vector, $\c_i$, that is fed to the decoder RNN is computed using the alignment, $\alphab_i$, to produce a weighted average of encoder states.
The decoder is fed both the context vector and the current attention RNN state, and an output function produces the decoder output, $\y_i$, from the decoder RNN state, $\d_i$.
\begin{align}
    \{\h_j\}_{j=1}^L &= \textrm{Encoder}(\{\x_j\}_{j=1}^L) \\
    \s_i &= \textrm{RNN}_\textrm{Att}(\s_{i-1}, \c_{i-1}, \y_{i-1}) \label{eq:attention_rnn} \\
    \alphab_i &= \textrm{Attention}(\s_i, ~\dots)
    &\c_i &= \sum_j \alpha_{i,j} \h_j \label{eq:attention_fn_context_vector}\\
    \d_i &= \textrm{RNN}_\textrm{Dec}(\d_{i-1}, \c_i, \s_i)
    &\y_i &= f_\textrm{o}(\d_i)
\end{align}

\subsection{GMM-Based Mechanisms}
\label{subsec:gmm_mechanisms}

An early sequence-to-sequence attention mechanism was proposed by Graves in \cite{Graves:2013ua}.
This approach is a purely location-based mechanism that uses an unnormalized mixture of $K$ Gaussians to produce the attention weights, $\alphab_i$, for each encoder state.
The general form of this type of attention is shown in \eqref{eq:gmm_attention}, where $\wb_i$, $\Zb_i$, $\Db_i$, and $\sigmab_i$ are computed from the attention RNN state.
The mean of each Gaussian component is computed using the recurrence relation in \eqref{eq:monotonic_gmm}, which makes the mechanism location-relative and potentially monotonic if $\Db_i$ is constrained to be positive.
\begin{align}
    \alpha_{i,j} &= \sum_{k=1}^K \frac{w_{i,k}}{Z_{i,k}} 
    \exp \left(-\frac{(j-\mu_{i,k})^2}{2(\sigma_{i,k})^2}\right) \label{eq:gmm_attention} \\
    \mub_i &= \mub_{i-1} + \Db_i \label{eq:monotonic_gmm}
\end{align}
In order to compute the mixture parameters, intermediate parameters ($\wh_i,\Dh_i,\sigmah_i$) are first computed using the MLP in \eqref{eq:gmm_mlp} and then converted to the final parameters using the expressions in Table~\ref{tab:gmm_attention}.
\begin{align}
    (\wh_i,\Dh_i,\sigmah_i) &= V \tanh(W\s_i + b) \label{eq:gmm_mlp}
\end{align}
The version 0 (V0) row in Table~\ref{tab:gmm_attention} corresponds to the original mechanism proposed in \cite{Graves:2013ua}.
V1 adds normalization of the mixture weights and components and uses the exponential function to compute the mean offset and variance.
V2 uses the softplus function to compute the mean offset and standard deviation.

Another modification we test is the addition of initial biases to the intermediate parameters $\Dh_i$ and $\sigmah_i$ in order to encourage the final parameters $\Db_i$ and $\sigmab_i$ to take on useful values at initialization.
In our experiments, we test versions of V1 and V2 GMM attention that use biases that target a value of $\Db_i=1$ for the initial forward movement and $\sigmab_i=10$ for the initial standard deviation (taking into account the different nonlinearities used to compute the parameters).
\begin{table}[htb]
    \caption{Conversion of intermediate parameters computed in \eqref{eq:gmm_mlp} to final mixture parameters for the three tested GMM-based attention mechanisms. $\softmax(\cdot)$ is the softmax function, while $\softplus(\cdot)$ is the softplus function.}
    \label{tab:gmm_attention}
    \begin{tabular}{lllll}
        \toprule
                                & $\Zb_i$                  & $\wb_i$           & $\Db_i$            & $\sigmab_i$\\
        \midrule
        V0~\cite{Graves:2013ua} & $\bm{1}$                 & $\e{\wh_i}$       & $\e{\Dh_i}$        & $\sqrt{\e{-\sigmah_i} / 2}$\\
        V1                      & $\sqrt{2\pi\sigmab_i^2}$ & $\softmax(\wh_i)$ & $\e{\Dh_i}$        & $\sqrt{\e{\sigmah_i}}$\\
        V2                      & $\sqrt{2\pi\sigmab_i^2}$ & $\softmax(\wh_i)$ & $\softplus(\Dh_i)$ & $\softplus(\sigmah_i)$\\
        \bottomrule
    \end{tabular}
\end{table}

\subsection{Additive Energy-Based Mechanisms}
\label{subsec:energy_based_mechanisms}

A separate family of attention mechanisms use an MLP to compute attention energies, $\eb_i$, that are converted to attention weights, $\alphab_i$, using the softmax function.
This family includes the content-based mechanism introduced in \cite{Bahdanau:2014vz} and the hybrid location-sensitive mechanism from \cite{Chorowski:2015uh}.
A generalized formulation of this family is shown in \eqref{eq:energy_based_mlp}.
\begin{align}
    e_{i,j} &= \bm{v}^\intercal \tanh(W\s_i + V\h_j + U\f_{i,j} + T\g_{i,j} + \b) + p_{i,j} \label{eq:energy_based_mlp} \\
    \alphab_i &= \softmax(\eb_i) \\
    \f_i &= \mathcal{F} * \alphab_{i-1} \label{eq:static_filters} \\
    \g_i &= \mcG(\s_i) * \alphab_{i-1}, \quad
    \mcG(\s_i) = V_\mcG \tanh(W_\mcG \s_i + \b_\mcG) \label{eq:dynamic_filters} \\
    \p_i &= \log(\mathcal{P} * \alphab_{i-1}) \label{eq:prior_filter}
\end{align}
Here we see the content-based terms, $W\s_i$ and $V\h_j$, that represent query/key comparisons and the location-sensitive term, $U\f_{i,j}$, that uses convolutional features computed from the previous attention weights as in \eqref{eq:static_filters}~\cite{Chorowski:2015uh}.
Also present are two new terms, $T\g_{i,j}$ and $p_{i,j}$, that are unique to our proposed Dynamic Convolution Attention.
The $T\g_{i,j}$ term is very similar to $U\f_{i,j}$ except that it uses dynamic filters that are computed from the current attention RNN state as in \eqref{eq:dynamic_filters}.
The $p_{i,j}$ term is the output of a fixed prior filter that biases the mechanism to favor certain types of alignment.
Table~\ref{tab:energy_based_attention} shows which of the terms are present in the three energy-based mechanisms we compare in this paper.
\begin{table}[htb]
    \caption{The terms from \eqref{eq:energy_based_mlp} that are present in each of the three energy-based attention mechanisms we test.}
    \label{tab:energy_based_attention}
    \begin{tabular}{lccccc}
        \toprule
                            & $W\s_i$ & $V\h_j$ & $U\f_{i,j}$ & $T\g_{i,j}$ & $p_{i,j}$ \\
        \midrule
        Content-Based~\cite{Bahdanau:2014vz}       & \cmark  & \cmark  & \xmark      & \xmark      & \xmark \\
        Location-Sensitive~\cite{Chorowski:2015uh}  & \cmark  & \cmark  & \cmark      & \xmark      & \xmark \\
        Dynamic Convolution & \xmark  & \xmark  & \cmark      & \cmark      & \cmark \\
        \bottomrule
    \end{tabular}
\end{table}

\subsection{Dynamic Convolution Attention}
\label{subsec:DCA}
In designing Dynamic Convolution Attention (DCA), we were motivated by location-relative mechanisms like GMM attention, but desired fully normalized attention weights.
Despite the fact that GMM attention V1 and V2 use normalized mixture weights and components, the attention weights still end up unnormalized because they are sampled from a continuous probability density function.
This can lead to occasional spikes or dropouts in the alignment, and 
attempting to directly normalize GMM attention weights results in unstable training.
Attention normalization isn't a significant problem in fine-grained output-to-text alignment, but becomes more of an issue for coarser-grained alignment tasks where the attention window needs to gradually move to the next index (for example in variable-length prosody transfer applications~\cite{Lee:2018uo}).
Because DCA is in the energy-based attention family, it is normalized by default and should work well for a variety of monotonic alignment tasks.

Another issue with GMM attention is that because it uses a mixture of distributions with infinite support, it isn't necessarily monotonic. 
At any time, the mechanism could choose to emphasize a component whose mean is at an earlier point in the sequence, or it could expand the variance of a component to look backward in time, potentially hurting alignment stability.

To address monotonicity issues, we make modifications to the hybrid location-sensitive mechanism.
First we remove the content-based terms, $W\s_i$ and $W\h_i$, which prevents the alignment from moving backward due to a query/key match at a past timestep.
Doing this prevents the mechanism from adjusting its alignment trajectory as it is only left with a set of static filters, $U\f_{i,j}$, that learn to bias the alignment to move forward by a certain fixed amount.
To remedy this, we add a set of learned \emph{dynamic} filters, $T\g_{i,j}$, that are computed from the attention RNN state as in \eqref{eq:dynamic_filters}.
These filters serve to dynamically adjust the alignment relative to the alignment at the previous step.

In order to prevent the dynamic filters from moving things backward, we use a single fixed prior filter to bias the alignment toward short forward steps.
Unlike the static and dynamic filters, the prior filter is a causal filter that only allows forward progression of the alignment.
In order to enforce the monotonicity constraint, the output of the filter is converted to the logit domain via the log function before being added to the energy function in \eqref{eq:energy_based_mlp} (we also floor the prior logits at $-10^6$ to prevent underflow).

\begin{figure}[htb]
    \centerline{\includegraphics[width=1.0\linewidth]{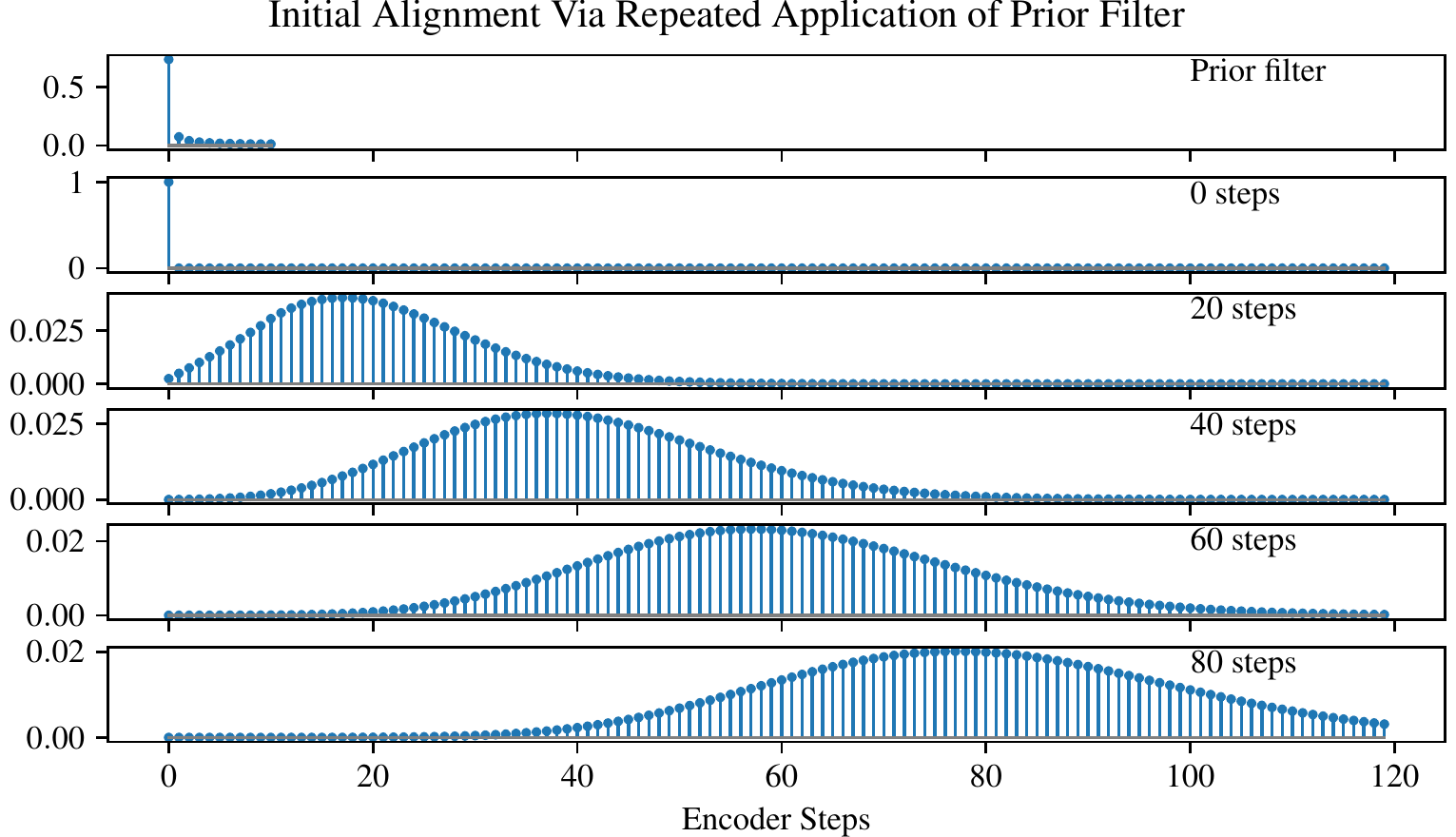}}
    \caption{
        Initial alignment encouraged by the prior filter (ignoring the contribution of other term in \eqref{eq:energy_based_mlp}).
        The attention weights are shown every 20 decoders steps with the prior filter itself shown at the top.
    }
    \label{fig:prior_alignment}
\end{figure}

We set the taps of the prior filter using values from the beta-binomial distribution, which is a two-parameter discrete distribution with finite support.
\begin{align}
    p(k) &= \binom{n}{k} \frac{\textrm{B}(k+\alpha, n-k+\beta)}{\textrm{B}(\alpha,\beta)}, \quad
    k \in \{0,\ldots,n\}
\end{align}
    where $\textrm{B}(\cdot)$ is the beta function.
For our experiments we use the parameters $\alpha=0.1$ and $\beta=0.9$ to set the taps on a length-11 prior filter ($n=10$).
Repeated application of the prior filter encourages an average forward movement of 1 encoder step per decoder step ($\mathbb{E}[k] = \alpha n/(\alpha+\beta)$) with the uncertainty in the prior alignment increasing after each step.
The prior parameters could be tailored to reflect the phonemic rate of each dataset in order to optimize alignment speed during training, but for simplicity we use the same values for all experiments.
Figure~\ref{fig:prior_alignment} shows the prior filter along with the alignment weights every 20 decoder steps when ignoring the contribution from other terms in \eqref{eq:energy_based_mlp}.

\section{Experiments}
\label{sec:experiments}

\subsection{Experiment Setup}
\label{subsec:experiment_setup}

In our experiments we compare the GMM and additive energy-based families of attention mechanisms enumerated in Tables \ref{tab:gmm_attention} and \ref{tab:energy_based_attention}.
We use the Tacotron architecture described in Section~\ref{subsec:basic_setup}
and only vary the attention function used to compute the attention weights, $\alphab_i$.
The decoder produces two 128-bin, 12.5ms-hop mel spectrogram frames per step.
We train each model using the Adam optimizer for 300,000 steps with a gradient clipping threshold of 5 and a batch size of 256, spread across 32 Google Cloud TPU cores.
We use an initial learning rate of $10^{-3}$ that is reduced to $5\times 10^{-4}$, $3\times 10^{-4}$, $10^{-4}$, and $5\times 10^{-5}$ at 50k, 100k, 150k, and 200k steps, respectively.
To convert the mel spectrograms produced by the models into audio samples, we use a separately-trained WaveRNN~\cite{Kalchbrenner:2018wr} for each speaker.

For all attention mechanisms, we use a size of 128 for all tanh hidden layers.
For the GMM mechanisms, we use $K=5$ mixture components.
For location-sensitive attention (LSA), we use 32 static filters, each of length 31.
For DCA, we use 8 static filters and 8 dynamic filters (all of length 21), and a length-11 causal prior filter as described in Section~\ref{subsec:DCA}.

We run experiments using two different single-speaker datasets. 
The first (which we refer to as the \emph{Lessac} dataset) comprises audiobook recordings from Catherine Byers, the speaker from the 2013 Blizzard Challenge.  
For this dataset, we train on a 49,852-utterance (37-hour) subset, consisting of utterances up to 5 seconds long, and evaluate on a separate 935-utterance subset.
The second is the LJ Speech dataset~\cite{ljspeech17}, a public dataset consisting of audiobook recordings that are segmented into utterances of up to 10 seconds. We train on a 12,764-utterance subset (23 hours) and evaluate on a separate 130-utterance subset.

\subsection{Alignment Speed and Consistency}
\label{subsec:alignment_speed}
To test the alignment speed and consistency of the various mechanisms, we run 10 identical trials of 10,000 training steps and plot the MCD-DTW between a ground truth holdout set and the output of the model during training.
The MCD-DTW is an objective similarity metric that uses dynamic time warping (DTW) to find the minimum mel cepstral distortion (MCD)~\cite{kubichek1993mel} between two sequences.
The faster a model is able to align with the text, the faster it will start producing reasonable spectrograms that produce a lower MCD-DTW.
\begin{figure}[htb]
    \begin{minipage}[b]{1.0\linewidth}
      \centering
      \centerline{\includegraphics[width=1.0\linewidth]{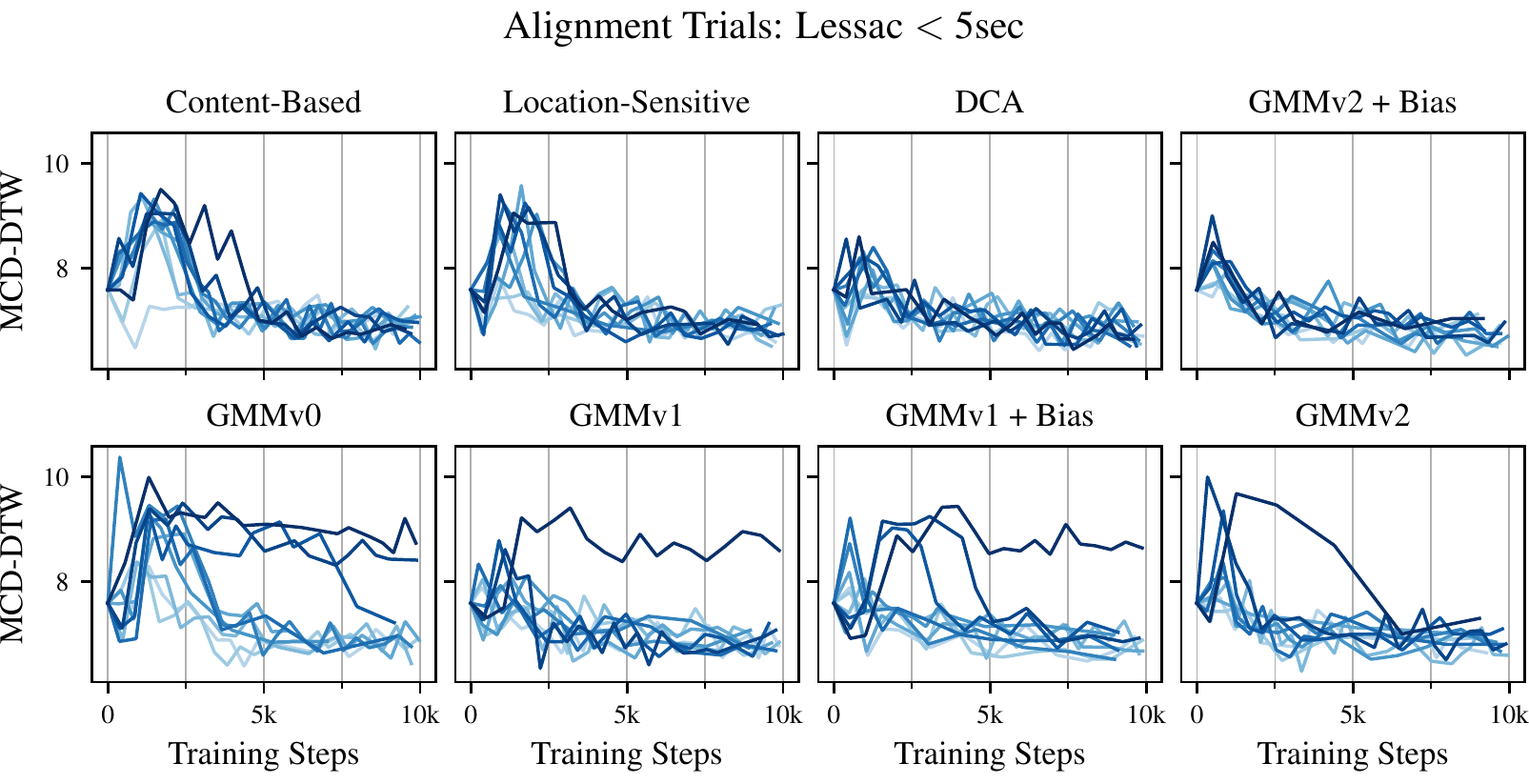}}
    \end{minipage}
    \begin{minipage}[b]{1.0\linewidth}
      \centering
      \centerline{\includegraphics[width=1.0\linewidth]{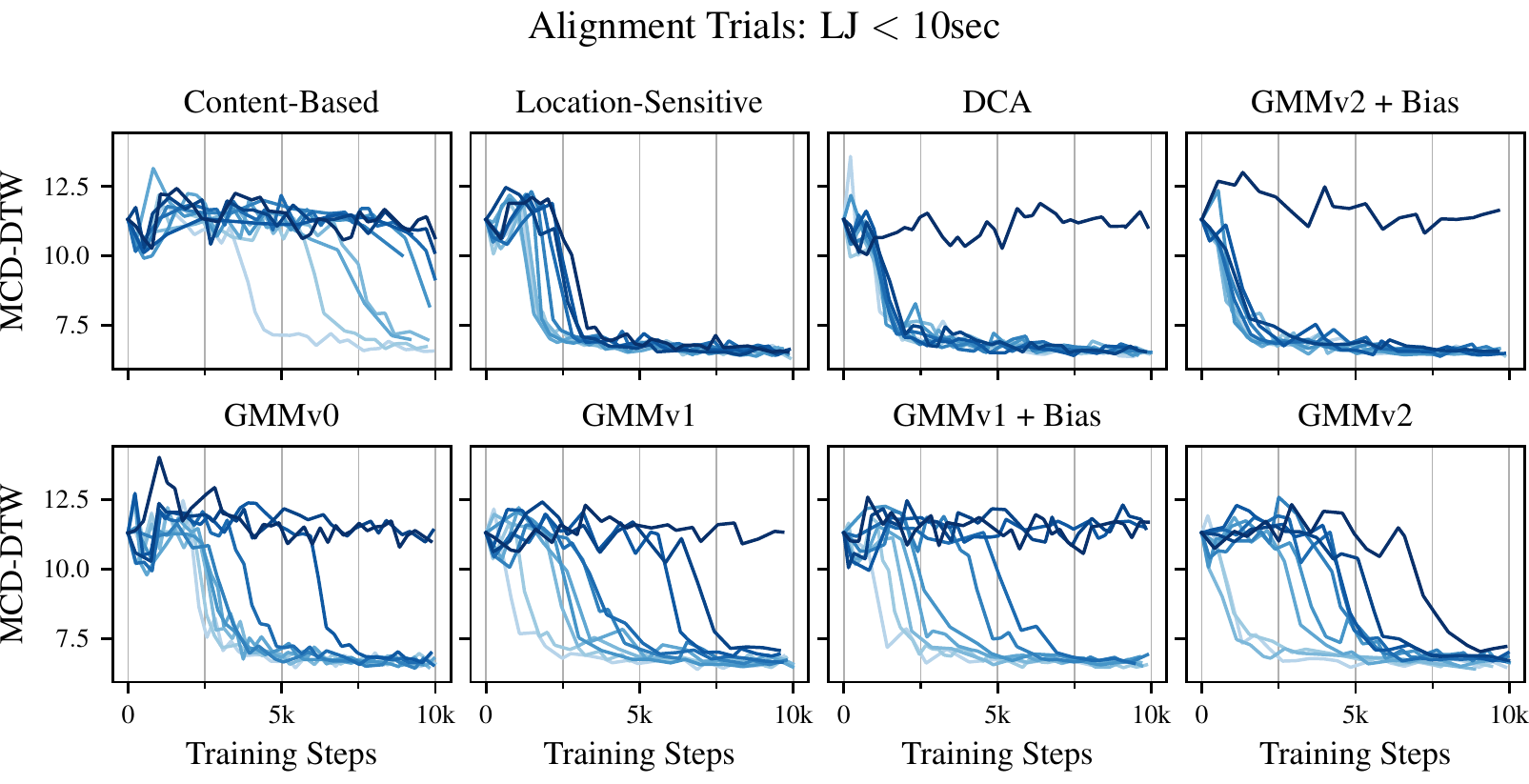}}
    \end{minipage}
    \caption{
        Alignment trials for 8 different mechanisms (10 runs each) trained on the Lessac (top) and LJ (bottom) datasets.
        The validation set MCD-DTW drops down after alignment has occurred.
    }
    \label{fig:alignment_trials}
\end{figure}

Figure~\ref{fig:alignment_trials} shows these trials for 8 different mechanisms for both the Lessac and LJ datasets. 
Content-based (CBA), location-sensitive (LSA), and DCA are the three energy-based mechanisms from Table~\ref{tab:energy_based_attention}, and the 3 GMM varieties are shown in Table~\ref{tab:gmm_attention}.
We also test the V1 and V2 GMM mechanisms with an initial parameter bias as described in Section~\ref{subsec:gmm_mechanisms} (abbreviated as GMMv1b and GMMv2b).

Looking at the plots for the Lessac dataset (top of Figure~\ref{fig:alignment_trials}), we see that the mechanisms on the top row (the energy-based family and GMMv2b) all align consistently, with DCA and GMMv2b aligning the fastest.
The GMM mechanisms on the bottom row don't fare as well, and while they typically align more often than not, there are a significant number of failures or cases of delayed alignment.
It's interesting to note that adding a bias to the GMMv1 mechanism actually hurts its consistency while adding a bias to GMMv2 helps it.

Looking at the plots for the LJ dataset at the bottom of Figure~\ref{fig:alignment_trials}, we first see that the dataset is more difficult in terms of alignment.
This is likely due to the higher maximum and average length of the utterances in the training data (most utterances in the LJ dataset are longer than 5 seconds) but could also be caused by an increased presence of intra-utterance pauses and overall lower audio quality.
Here, the top row doesn't fare as well: CBA has trouble aligning within the first 10k steps, while DCA and GMMv2b both fail to align once.
LSA succeeds on all 10 trials but tends to align more slowly than DCA and GMMv2b when they succeed.
With these consistency results in mind, we will only be testing the top row of mechanisms in subsequent evaluations.

\subsection{In-Domain Naturalness}
\label{subsec:naturalness}
We evaluate CBA, LSA, DCA, and GMMv2b using mean opinion score (MOS) naturalness judgments produced by a crowd-sourced pool of raters. 
Scores range from 1 to 5, with 5 representing ``completely natural speech''.
The Lessac and LJ models are evaluated on their respective test sets (hence in-domain), and the results are shown in Table~\ref{tab:mos}.
We see that for these utterances, the LSA, DCA, and GMMV2b mechanisms all produce equivalent scores around 4.3, while the content-based mechanism is a bit lower due to occasional catastrophic attention failures.
\begin{table}[htb]
    \caption{MOS naturalness results along with 95\% confidence intervals for the Lessac and LJ datasets.}
    \label{tab:mos}
    \centering
    \begin{tabular}{lcc}
    \toprule
                 & Lessac          & LJ              \\
    \midrule
    Content-Based      & 4.07 \PM 0.08 & 4.19 \PM 0.06 \\
    Location-Sensitive & 4.31 \PM 0.06 & 4.34 \PM 0.06 \\
    GMMv2b             & 4.32 \PM 0.06 & 4.29 \PM 0.06 \\
    DCA                & 4.31 \PM 0.06 & 4.33 \PM 0.06 \\
    Ground Truth       & 4.64 \PM 0.04 & 4.55 \PM 0.04 \\
    \bottomrule
    \end{tabular}    
\end{table}

\subsection{Generalization to Long Utterances}
\label{subsec:generalization_to_long_utterances}
Now we evaluate our models on long utterances taken from two chapters of the Harry Potter novels.
We use 1034 utterances that vary between 58 and 1648 characters (10 and 299 words).
Google Cloud Speech-To-Text\footnote{\url{https://cloud.google.com/speech-to-text}} is used to produce transcripts of the resulting audio output, and we compute the character errors rate (CER) between the produced transcripts and the target transcripts.
\begin{figure}[htb]
    \begin{minipage}[b]{1.0\linewidth}
      \centering
      \centerline{\includegraphics[width=1.0\linewidth]{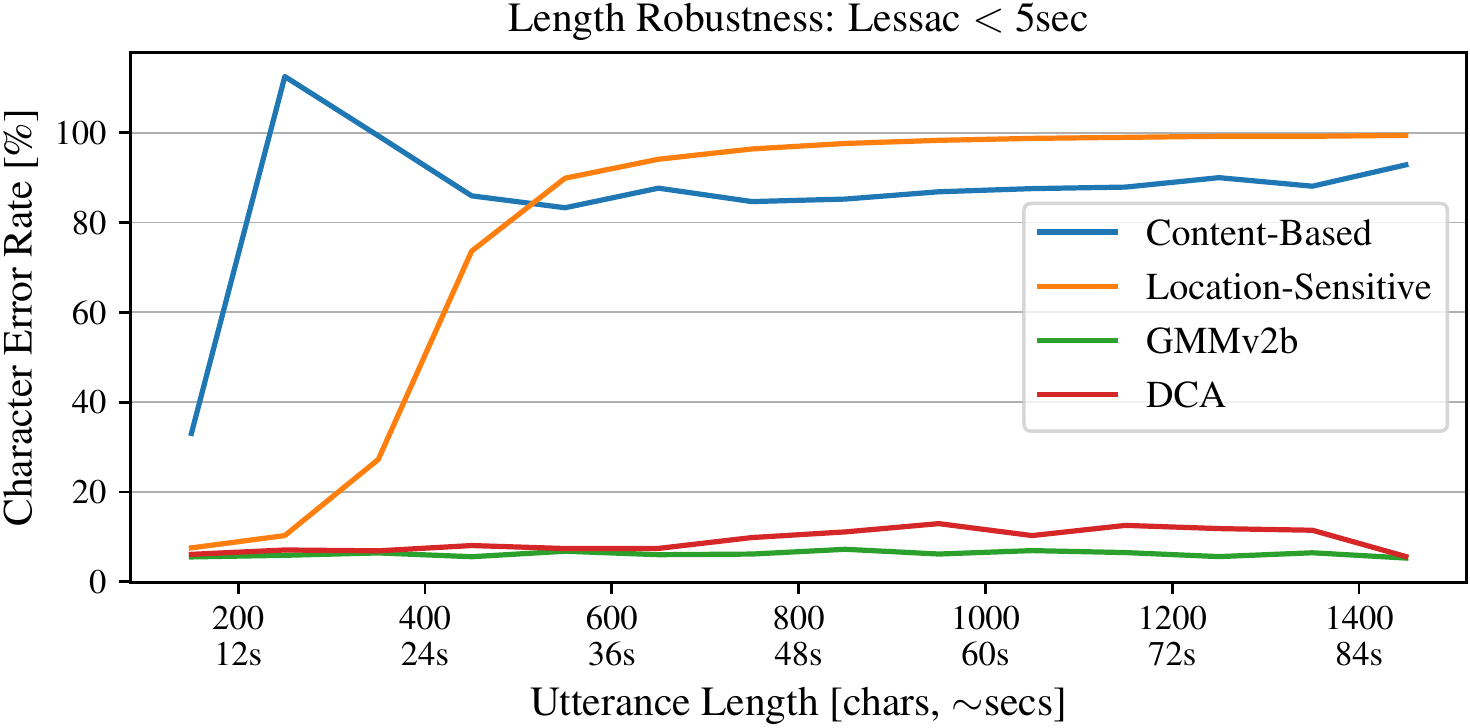}}
    \end{minipage}
    \begin{minipage}[b]{1.0\linewidth}
      \centering
      \centerline{\includegraphics[width=1.0\linewidth]{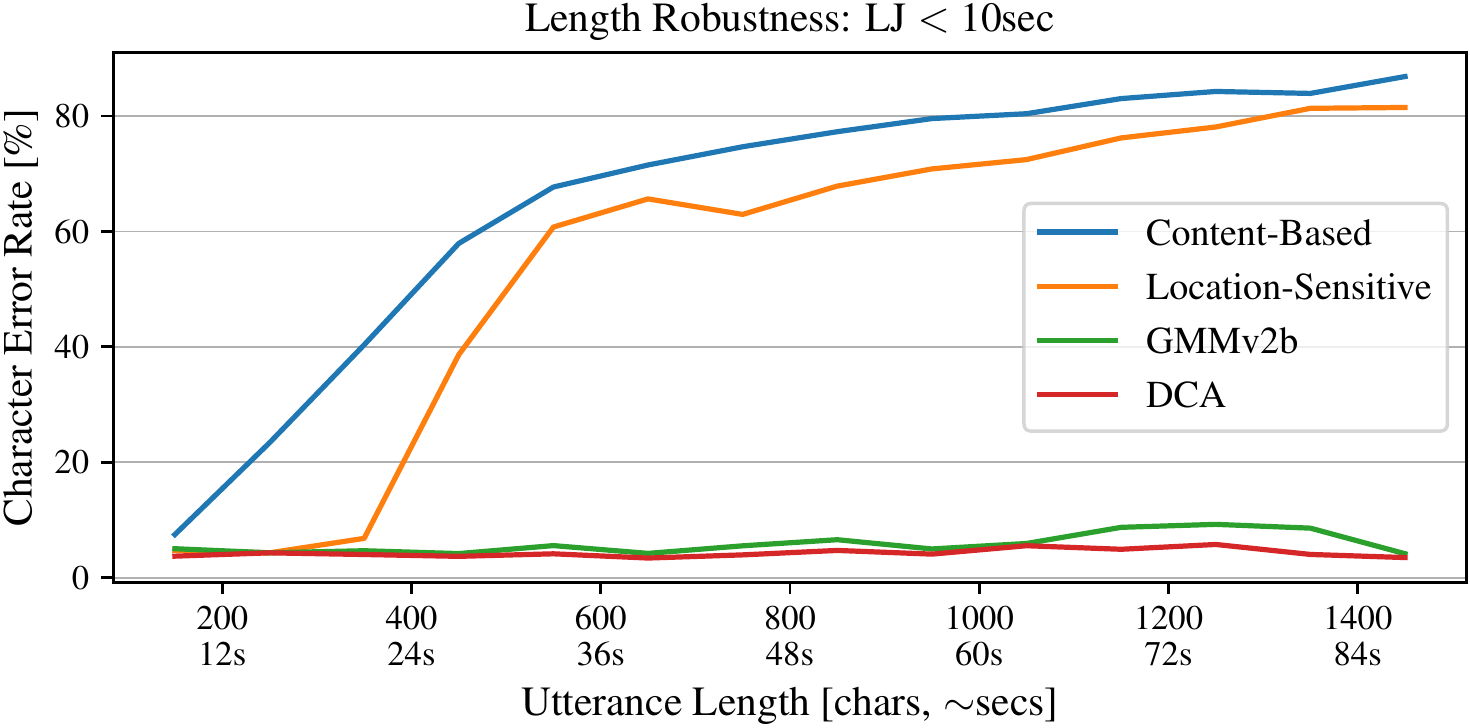}}
    \end{minipage}
    \caption{
        Utterance length robustness for models trained on the Lessac (top) and LJ (bottom) datasets.
    }
    \label{fig:asr}
\end{figure}
Figure~\ref{fig:asr} shows the CER results as a function of utterance length for the Lessac models (trained on up to 5 second utterances) and LJ models (trained on up to 10 second utterances).
The plots show that CBA fares the worst, with the CER shooting up when the test length exceeds the max training length. 
LSA shoots up soon after at around 3x the max training length, while
the two location-relative mechanisms, DCA and GMMv2b, are both able to generalize to the whole range of utterance lengths tested.

\section{Discussion}
\label{sec:discussion}
We have shown that Dynamic Convolution Attention (DCA) and V2 GMM attention with initial bias (GMMv2b) are able to generalize to utterances much longer than those seen during training, while preserving naturalness on shorter utterances.
This opens the door for synthesis of entire paragraphs or long sentences (e.g., for book or news reading applications), which can improve naturalness and continuity compared to synthesizing each sentence or clause separately and then stitching them together.

These two location-relative mechanisms are simple to implement and do not rely on dynamic programming to marginalize over alignments.
They also tend to align very quickly during training, which makes the occasional alignment failure easy to detect so training can be restarted.
In our alignment trials, despite being slower to align on average, LSA seemed to have an edge in terms of alignment consistency; however, we have noticed that slower alignment can sometimes lead to worse quality models, probably because the other model components are being optimized in an unaligned state for longer.

Compared to GMMv2b, DCA can more easily bound its receptive field (because its prior filter numerically disallows backward or excessive forward movement), which makes it easier to incorporate hard windowing optimizations in production.
Another advantage of DCA over GMM attention is that its attention weights are normalized, which helps to stabilize the alignment, especially for coarse-grained alignment tasks. 

For monotonic alignment tasks like TTS and speech recognition,
location-relative attention mechanisms have many advantages and warrant increased consideration and further study.
Supplemental materials, including audio examples, are available on the web\footnote{\url{\weblink}}.

\vfill\pagebreak
\bibliographystyle{IEEEbib}
\bibliography{refs}

\end{document}